\documentclass{article}
\usepackage[utf8]{inputenc}
 \usepackage[T1]{fontenc}
\usepackage{amsmath}
\usepackage{pifont}
\usepackage{bm}
 \usepackage{amssymb}
 \usepackage{amsthm}
\usepackage{parskip}
\usepackage[ruled, noend, linesnumbered]{algorithm2e}
\usepackage[colorlinks = true, pdfstartview = FitV, linkcolor = blue, citecolor = blue, urlcolor = blue, bookmarks = false]{hyperref}
\usepackage[authoryear, round]{natbib}
\usepackage{color,xcolor}
\usepackage{setspace} 
\usepackage{graphicx}
\usepackage{subcaption}
\usepackage{setspace}
\usepackage{xcolor}

\usepackage{caption}
\usepackage{tabularray}
\UseTblrLibrary{booktabs}

\usepackage{listings}
\usepackage{color} 

\definecolor{codegreen}{rgb}{0,0.6,0}
\definecolor{codegray}{rgb}{0.5,0.5,0.5}
\definecolor{codepurple}{rgb}{0.58,0,0.82}
\definecolor{backcolour}{rgb}{0.95,0.95,0.92}

\newcommand{\code}[1]{\colorbox{gray!20}{\texttt{#1}}}

\usepackage[margin=1.25
in]{geometry}
\newcommand{\m}[1]{\mathrm{#1} }

\renewcommand{\v}[1]{\boldsymbol{#1}}

\newcommand{\tp}{\tilde{p}}
\newcommand{\tq}{\tilde{q}}

\newcommand{\ignore}[1]{}

\newsavebox\IBoxA \newsavebox\IBoxB \newlength\IHeight
\newcommand\TwoFig[6]{
  \sbox\IBoxA{\includegraphics[width=0.45\textwidth]{#1}}
  \sbox\IBoxB{\includegraphics[width=0.45\textwidth]{#4}}%
  \ifdim\ht\IBoxA>\ht\IBoxB
    \setlength\IHeight{\ht\IBoxB}%
  \else\setlength\IHeight{\ht\IBoxA}\fi
  \begin{figure}[!htb]
  \minipage[t]{0.45\textwidth}\centering
  \includegraphics[height=\IHeight]{#1}
  \caption{#2}\label{#3}
  \endminipage\hfill
  \minipage[t]{0.45\textwidth}\centering
  \includegraphics[height=\IHeight]{#4}
  \caption{#5}\label{#6}
  \endminipage 
  \end{figure}%
}

\title{Federated Learning for Non-factorizable Models using Deep Generative Prior Approximations}
\author{Conor Hassan\footnote{\url{conorhassan.ai@gmail.com}} $\textsuperscript{ 1,2}$, Joshua J. Bon$\textsuperscript{ 3}$, Elizaveta Semenova$\textsuperscript{ 4}$, \\Antonietta Mira$\textsuperscript{ 5,6}$, Kerrie Mengersen$\textsuperscript{ 1,2}$ \\ 
{\small \textsuperscript{1} Centre for Data Science, Queensland University of Technology} \\ 
{\small \textsuperscript{2} School of Mathematical Sciences, Queensland University of Technology}  \\ 
{\small \textsuperscript{3} CEREMADE, Université Paris-Dauphine}  \\
{\small \textsuperscript{4} Imperial College London}  \\
{\small\textsuperscript{5} Euler Institute, Università della Svizzera italiana (USI)}\\
{\small\textsuperscript{6} Insubria University}
}

\date{}
\begin{document}
\maketitle
\begin{abstract}
Federated learning (FL) allows for collaborative model training across decentralized clients while preserving privacy by avoiding data sharing. However, current FL methods assume conditional independence between client models, limiting the use of priors that capture dependence, such as Gaussian processes (GPs). We introduce the \emph{S}tructured \emph{I}ndependence via deep \emph{G}enerative \emph{M}odel \emph{A}pproximation (SIGMA) prior which enables FL for non-factorizable models across clients, expanding the applicability of FL to fields such as spatial statistics, epidemiology, environmental science, and other domains where modeling dependencies is crucial. The SIGMA prior is a pre-trained deep generative model that approximates the desired prior and induces a specified conditional independence structure in the latent variables, creating an approximate model suitable for FL settings. We demonstrate the SIGMA prior's effectiveness on synthetic data and showcase its utility in a real-world example of FL for spatial data, using a conditional autoregressive prior to model spatial dependence across Australia. Our work enables new FL applications in domains where modeling dependent data is essential for accurate predictions and decision-making.
\end{abstract}

\section{Introduction}

Federated learning (FL) enables collaborative model estimation across multiple clients without needing centralized data aggregation \citep{mcmahan2017communication, kairouz2021advances}. FL addresses critical data privacy, security, and ownership concerns by enabling model training on decentralized data. However, a fundamental limitation of current FL algorithms is the assumption of strict conditional independence between the data from different clients \citep{kotelevskii2022fedpop, hassan2023federated}. This assumption implies that the local parameters of each client are independent given the global parameters, but this is often not true in real-world settings, such as spatio-temporal modeling or graph-structured data, where observations across clients exhibit complex dependencies. As a result, applying FL to these domains requires additional, suboptimal assumptions that limit the effectiveness and accuracy of the learned models.

To address this challenge, we introduce \emph{\textbf{S}tructured \textbf{I}ndependence via deep \textbf{G}enerative \textbf{M}odel \textbf{A}pproxi\-mation} (SIGMA) priors, a novel approach to enabling scalable FL algorithms for models with dependent data. The key idea is to train a hierarchical variational autoencoder (VAE) to approximate the prior distribution over client-specific parameters. The SIGMA prior learns a latent representation that captures dependencies between clients while providing a conditionally-independent structure, enabling the use of existing efficient FL algorithms.

We consider probabilistic models of the form
\begin{align} \label{eq:model_class}
\v \phi &\sim p(\v \phi), \\
\v \theta_j | \v \theta_{-j}, \v \phi &\sim p(\v \theta_{j} | \v \theta_{{-j}}, \v \phi), \hspace{0.3cm} j\in\{ 1, \ldots, J \},  \label{eq:difficult_density}\\
\v y_j | \v  \theta_{j}, \v \phi & \stackrel{\rm ind}{\sim}  p(\v y_j | \v \theta_j, \v \phi), \quad j \in \{1, \ldots, J\}, \label{eq:model_class_end}
\end{align}
where $\v \phi$ are global parameters shared across all $J$ clients, $\v \theta_{j}$ are local parameters specific to client $j$, $\v \theta_{{-j}}$ denotes local parameters from all clients except $j$, and $\v y_j$ is the data available to client $j$.
Existing FL algorithms, such as structured federated variational inference \citep[SFVI,][]{hassan2023federated} and FedSOUL \citep{kotelevskii2022fedpop}, restrict \eqref{eq:difficult_density} with a local conditional independence property, that is,
\begin{align}\label{eq:desired_factorization}
p(\v \theta_j | \v\theta_{-j}, \v\phi) = p(\v \theta_{j} | \v \phi), \quad j \in \{1, \ldots, J\}.
\end{align}

However, prior densities that do not satisfy \eqref{eq:desired_factorization} are common in statistics and machine learning. For example, Markov random field \citep[MRF,][]{besag1991bayesian, rue2009approximate} and Gaussian process (GP) priors \citep{bernardo1998regression, williams2006gaussian} are a common choice to capture spatial and temporal dependence between observations in applications such as image inpainting \citep{ruvzic2014context}, disease mapping \citep{lee2011comparison, riebler2016intuitive}, and graph neural networks \citep{hasanzadeh2020bayesian}.

\begin{figure}[ht!]
\centering
\includegraphics[width=0.95\textwidth]{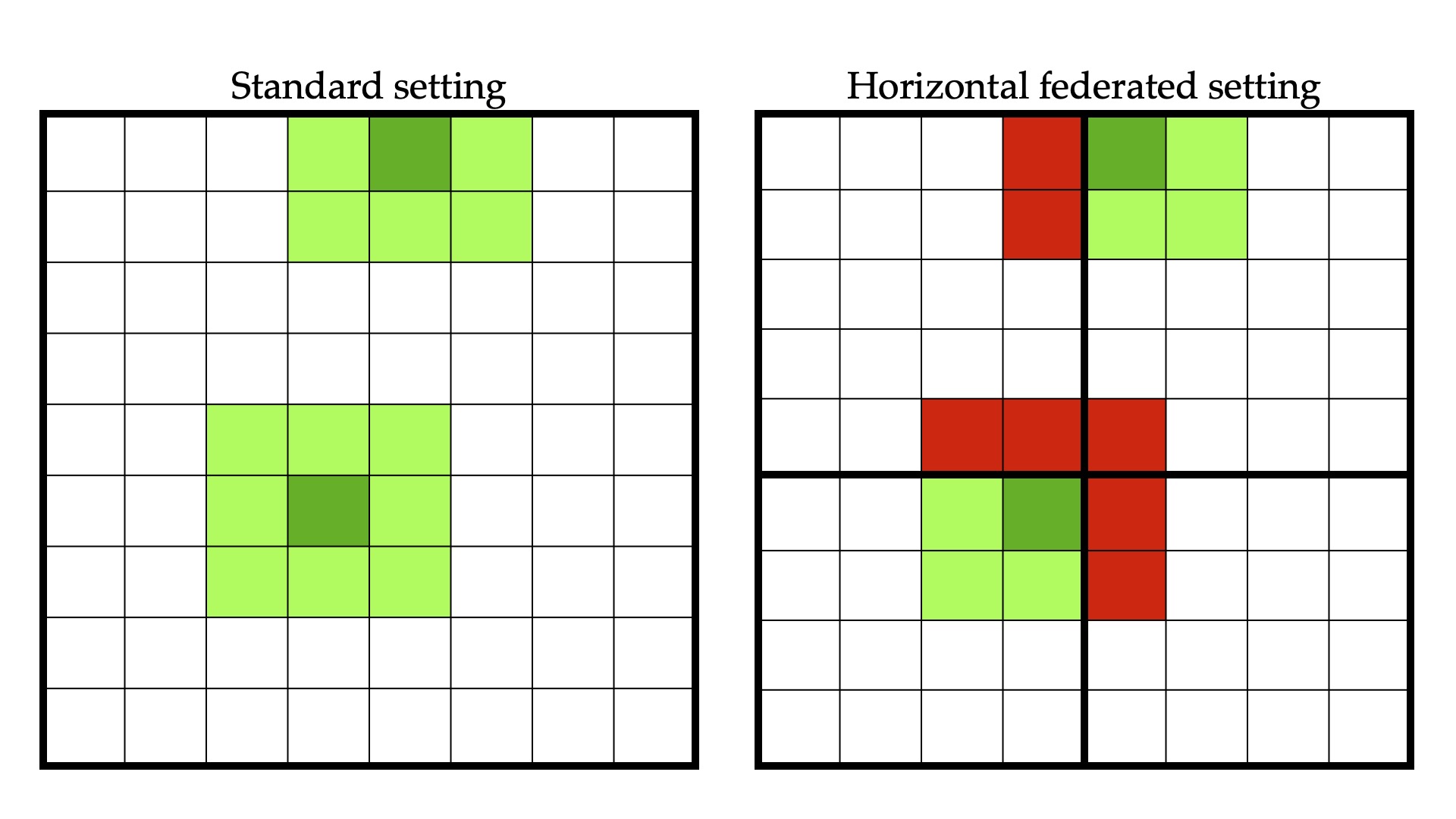}
\caption{The \emph{standard setting} on the left shows the dependency structure for a \emph{conditional autoregressive} (CAR) prior in a non-FL setting. On the right, the bold borders represent boundaries for different clients, and the red grids represent values we cannot evaluate when trying to infer, in an FL setting, the value at the dark-green grids.}
\label{fig:CAR_structure}
\end{figure}

Figure \ref{fig:CAR_structure} illustrates the challenge of applying such priors in a federated setting. The left panel shows the standard dependency structure for a conditional autoregressive (CAR) prior in a non-FL setting. In the FL setting (right panel), the bold borders represent boundaries between clients, and the red grids indicate values that cannot be evaluated when inferring the value at the dark-green grids due to the lack of access to data from other clients.

Our work aims to address this challenge by developing a novel approach that can capture the dependencies among local parameters while still enabling efficient and privacy-preserving learning in the FL setting. 
The SIGMA prior enables the use of computationally scalable FL algorithms for models with assumed dependencies between observations captured by the prior structure. 

The remainder of the paper is structured as follows. Section \ref{section:federated_inference} provides background on the model class of interest, SFVI, and VAEs. Section \ref{section:methods} introduces the SIGMA prior and our proposed methodology. Section \ref{section:numerical_examples} presents the results of our experiments on synthetic and real-world applications. Finally, Section \ref{section:discussion} concludes the paper with a discussion of the limitations and future directions for this work.

\section{Preliminaries}\label{section:federated_inference}
This section lays the foundation for our work by providing an overview of the techniques used in our approach, namely structured federated variational inference \citep[SFVI,][]{hassan2023federated}, \citep[VAEs,][]{kingma2013auto, rezende2014stochastic}, and PriorCVAE \citep{semenova2023priorcvae}.

We denote the $p\times p$ identity matrix with $I_p$. For $\v z \in \mathbb{R}^p$ we say $\v z \sim \mathcal{N}_p(\v\mu,\v\sigma)$ for  $p$-vectors $\v\mu$ and $\v\sigma$ if $z_i$ is normally distributed with mean and standard deviation 
($\mu_i$, $\sigma_i$) 
and independent of the other elements $\v z_{-i}$ for each $i 
\in \{1,\ldots,p\}$. If $\v z \sim \mathcal{N}_p(\v\mu,\Sigma)$ and $\Sigma$ is a $p\times p$ matrix, then $\v z$ has a multivariate normal distribution with mean $\v\mu$ and covariance matrix $\Sigma$.

\subsection{Structured Federated Variational Inference}\label{section:SFVI}
Variational inference \citep[VI,][]{blei2017variational, zhang2018advances} is an algorithmic technique that approximates the posterior distribution $p(\v\theta|\v y)$, which is often intractable for analytical or direct computation due to the complexity of the model. The key idea of VI is to introduce a parameterized distribution $q_{\v\lambda}(\v\theta) \in \mathcal{Q}$ called the \emph{variational approximation}, that aims to approximate the posterior distribution. The variational parameters $\v\lambda$ are optimized to find the best approximation from a fixed \emph{variational family} of distributions, $\mathcal{Q}$.

The objective in the VI framework is to maximize the \emph{evidence lower bound} (ELBO), defined as:
\begin{align}\label{eqn:elbo}
\mathcal{L}(\v\lambda) := \mathbb{E}_{\v \theta \sim q_{\v\lambda}}[\log p(\v \theta, \v y)-\log q_{\v\lambda}(\v \theta)],
\end{align}
which acts as a surrogate objective function, providing a lower bound to the log marginal likelihood. Maximizing the ELBO is equivalent to minimizing the Kullback-Leibler (KL) divergence between the variational approximation $q_{\v\lambda}(\v\theta)$ and the posterior distribution $p(\v\theta|\v y)$. When the variational approximation matches the true posterior, the ELBO equals the log marginal likelihood.

Gradient-based optimization algorithms \citep{ruder2017overview} are commonly used to maximize the ELBO, relying on unbiased gradient estimators. A significant advancement in stochastic VI \citep{hoffman2013stochastic} is the \emph{reparameterization gradient estimator} \citep{titsias2014doubly, kingma2013auto, rezende2014stochastic}, which allows for an efficient and low-variance estimate of the ELBO. The key idea is to express a sample of $\v\theta$ drawn from $q_{\v\lambda}(\v\theta)$ as a deterministic function of $\v\lambda$ by transforming random noise $\v\epsilon$: $\v\theta = f_{\v\lambda}(\v \epsilon)$. The noise variable $\v\epsilon$ is drawn from a distribution $q_{\v\epsilon}$, which is independent of $\v\lambda$. This reparameterization enables the computation of low-variance gradients, which is crucial for efficient optimization in VI. A resulting single-sample unbiased Monte Carlo estimator of the desired gradient vector is the \emph{sticking-the-landing} (STL) estimator \citep{roeder2017sticking},
\begin{align}\label{eqn:stl_estimator}
\widehat{\nabla}_{\v\lambda}\mathcal{L} := {\frac{\partial f_{\v\lambda}(\v\epsilon)}{\partial\v\lambda}}^\top \nabla_{\v \theta}[\log p(\v \theta, \v y) - \log q_{\v\lambda}(\v \theta)].
\end{align}

Structured federated variational inference \citep[SFVI,][]{hassan2023federated} extends VI to the FL setting, focusing on hierarchical Bayesian models of the form, 
\begin{align*}
p(\v y, \v \phi, \v\theta) = p(\v \phi) \prod_{j=1}^J p(\v \theta_j|\v \phi)p(\v y_j | \v\theta_j , \v\phi).
\end{align*}
where 
the local parameters $\v\theta_j$ only appear in the likelihood component for the $j$-th client, and the sets of local parameters are assumed to be conditionally independent given the global parameters. To enable efficient and privacy-preserving learning in the federated setting, SFVI employs a factorized structured variational approximation \citep{hoffman2015stochastic, ranganath2016hierarchical, tan2020conditionally} of the form:
\begin{align*}
q(\v \phi, \v\theta) = q_{\v\varphi_G}(\v \phi)\prod_{j=1}^J q_{\v\varphi_{L_j}}(\v \theta_j|\v \phi),
\end{align*}
where $q_{\v\varphi_G}(\v \phi)$ is a Gaussian distribution parameterized by $\v \varphi_G$, and each $q_{\v\varphi_{L_j}}(\v \theta_j|\v \phi )$ is a Gaussian distribution parameterized by $\v \varphi_{L_j}$, potentially including dependencies on $\v \phi$. This factorization allows for local optimization and privacy-preserving updates, as the local variational parameters $\v\varphi_{L_j}$ can be updated independently by each client without sharing them with the server. SFVI leverages the factorized structure of the variational approximation to decompose the ELBO into a sum of local ELBOs, each of which can be optimized locally by a client. Two algorithms, SFVI and SFVI-Avg, are derived for combinations of models and variational approximations that share this form. Both algorithms enable parallel and private updates of the local variational parameters $\v\varphi_{L_j}$ for each client, requiring only the communication of gradient information concerning the global variational parameters $\v\varphi_G$ to the server at each iteration.

In the numerical examples presented in this work, we focus on the SFVI algorithm, as it returns the exact estimates that fit the equivalent variational approximation on the full dataset in the non-FL setting.
While SFVI provides a principled and efficient approach to learning hierarchical Bayesian models in federated settings, it assumes that the local parameters are conditionally independent given the global parameters. This assumption may limit the applicability of SFVI to models with dependent local parameters, which are common in real-world applications. In Section \ref{section:methods}, we introduce the SIGMA prior, a novel approach to relaxing this assumption and enabling the application of SFVI to a broader class of models with dependent local parameters. By capturing the dependencies among local parameters while still allowing for efficient and privacy-preserving learning, the SIGMA prior has the potential to significantly expand the scope and impact of FL in various domains.

\subsection{Variational Autoencoders and PriorCVAE}\label{section:prelim_vae}
\subsubsection{Variational Autoencoders}
\emph{Variational autoencoders} \citep[VAE,][]{kingma2013auto, rezende2014stochastic} are a class of latent variable models that use variational approximation techniques to facilitate estimation of the latent space~$\v z$. 
A VAE consists of a generative model,~$\tp_{\v\psi}(\v\theta, \v z) = \tp_{\v\psi}(\v\theta\vert\v z)\tp(\v z)$, coupled with $\tq_{\v\varphi}(\v z \vert \v\theta)$, an amortised variational approximation of~$\tp_{\v\psi}(\v z\vert\v\theta)$, the posterior distribution implied by the generative model. The use of the variational approximation facilitates learning the components of the VAE. We will refer to the generative model of the VAE simply as the \emph{generative model} and use~$\tp$ to distinguish this model from the true model that it is approximating.

The task of a VAE is to approximate some distribution $p(\v\theta)$ using a latent variable representation $\tp_{\v\psi}(\v\theta) = \int \tp_{\v\psi}(\v\theta\vert\v z)\tp(\v z) \text{d} z$ given by the chosen generative model. As in VI (see Section~\ref{section:SFVI}), the objective function employed by VAEs is the appropriate average ELBO,
\begin{equation}\label{eq:elbovae}
    \mathcal{L}(\v\psi,\v\varphi) = \mathbb{E}_{\v \theta \sim p} \mathbb{E}_{\v z \sim \tq_{\v\varphi}(\v z \vert \v\theta)}[\log \tp_{\v \psi}(\v\theta , \v z) - \log \tq_{\v\varphi}(\v z \vert \v\theta)],
\end{equation}
where the first average is taken with respect to the true data generating process $\v \theta \sim p(\v\theta)$. Note that replacing the approximation $\tq_{\v\varphi}(\v z \vert \v\theta)$ with the correct $\tp_{\v\psi}(\v z \vert \v\theta)$ simplifies \eqref{eq:elbovae} to $\mathcal{L}(\v\psi,\v\varphi) = \mathbb{E}_{\v \theta \sim p}[\log \tp_{\v\psi}(\v\theta)]$ which is the full objective function, as opposed to the ELBO. As such, we can view the variational posterior, $\tq_{\v\varphi}(\v z \vert \v\theta)$, as an approximation to avoid the difficulties in evaluating $\tp_{\v\psi}(\v z \vert \v\theta)$.

Deep latent Gaussian models \citep[DLGMs,][]{rezende2014stochastic} are a popular type of generative model commonly used within a VAE. The generative model for a single-layer DLGM consists of a latent variable $\v z \sim \mathcal{N}_p(\v 0, I_p)$ and conditional variable $\v \theta | \v z\sim\mathcal{N}_d(\v\mu_{\v\psi}(\v z), \v\sigma_{\v\psi}(\v z))$ where $\mathcal{D}_{\v \psi} = (\v\mu_{\v \psi},\log \v\sigma_{\v \psi})$ 
are obtained via the joint output of a neural network termed the \emph{decoder}, parameterized by~$\v\psi$. The variational approximation $\tq_{\v\varphi}(\v z \vert \v\theta)$ is constructed from an \emph{encoder},~$\mathcal{E}_{\v \varphi} = (\v\mu_{\v \varphi},\log \v\sigma_{\v\varphi})$ such that $\v z | \v \theta \sim \mathcal{N}_p(\v\mu_{\v\varphi}(\v \theta), \v\sigma_{\v\varphi}(\v \theta))$. The combination of using a decoder in the generative model and using an encoder to parameterize the variational approximation results in a probabilistic analog to the autoencoder \citep{tschannen2018recent}.


VAEs were originally designed to construct a generative model for finite observed data, say $\{x_i\}_{i=1}^{n}$. In this case, $p(\v\theta)$ is an empirical distribution, $p(\v\theta) = \sum_{i=1}^n \delta_{x_i}(\v\theta)$, and the outer expectation of \eqref{eq:elbovae} becomes an empirical average over the data $\{x_i\}_{i=1}^{n}$. Estimation then typically involves stochastic gradient descent \citep[SGD,][]{ruder2017overview} to handle the inner expectation of \eqref{eq:elbovae}. Mini-batching of the data \citep{hoffman2013stochastic} is also frequently used and provides an unbiased estimate of the outer expectation of \eqref{eq:elbovae}. VAEs can also be used when $p(\v\theta)$ is a non-empirical distribution that can be sampled from (e.g. continuous). In this case, SGD is used to handle both expectations. We present this approach for the case of prior approximation next.

\subsubsection{PriorVAE and PriorCVAE}\label{subsection:priorCVAE}
PriorVAE \citep{semenova2022priorvae} adopts a VAE approach to learn an approximate prior distribution, which can be used in place of the desired prior.
Such an approximation aims to reduce the dimensionality or difficulty of an inferential task when the prior is the computational bottleneck because, for example, it is costly to evaluate. PriorVAE uses a decoder $\mathcal{D}_{\v\psi} = (\v\mu_{\v\psi}, \log\gamma)$ with fixed variance $\gamma^2$ and optimizes \eqref{eq:elbovae} using SGD with draws $\v\theta \sim p(\v\theta)$ from the desired prior distribution.

The approximate prior is constructed from the VAE by taking~$\v\theta = \v\mu_{\v\psi}(\v z)$ with~$\v z \sim N_p(\v 0,I_p)$, or more generally $\v z \sim \tp(\v z)$, where $\v\mu_{\v\psi}(\v z)$ is the mean output of the decoder. Using the approximate prior changes the posterior distribution from $\pi(\v\theta \vert \v x) \propto p(\v x\vert \v \theta)p(\v\theta)$ over $\v\theta$ to the approximate posterior $\tilde{\pi}(\v z \vert \v x) \propto p(\v x\vert \v \mu_{\v\psi}(\v z))\tp(\v z)$ over $\v z$. The $\v\theta$ implied by the approximate posterior is recovered using~$\v\theta=\v\mu_{\v\psi}(\v z)$. One can choose the dimension of the latent variables $p \ll d$, ensuring that the effective dimension of the approximate posterior is much less than the original posterior. Large computational benefits during inference were observed by \citet{semenova2022priorvae}, who motivate their work from this perspective.

PriorCVAE \citep{semenova2023priorcvae} extends PriorVAE to approximate conditional priors of the form $p(\v\theta, \v\phi) = p(\v\theta|\v\phi)p(\v\phi)$ where~$\v\phi$ is a random variable that forms a hierarchy in the prior structure. Rather than approximating $p(\v\theta, \v\phi)$ entirely, PriorCVAE only approximates the conditional prior $p(\v\theta|\v\phi)$, thus retaining $p(\v\phi)$ in both the original and approximate prior. With this extension, the ELBO \eqref{eq:elbovae} becomes
\begin{equation}\label{eq:elbocvae}
    \mathcal{L}(\v\psi,\v\varphi) = \mathbb{E}_{(\v\theta, \v\phi) \sim p} \mathbb{E}_{\v z \sim \tq_{\v\varphi}(\v z \vert \v\theta, \v\phi)}[\log \tp_{\v \psi}(\v\theta , \v z \vert \v\phi) - \log \tq_{\v\varphi}(\v z \vert \v\theta, \v\phi)].
\end{equation}
For the remainder of the paper, we will only consider the conditional variant of VAEs. To illustrate the PriorCVAE approximation we present a toy Gaussian Process regression example in Table~\ref{tab:GPexample}.
\begin{table}
    \centering
    \begin{tabular}{ll} \hline
        True model & PriorCVAE model  \\ \hline
        $\sigma \sim p(\sigma)$ & $\sigma \sim p(\sigma)$ \\
        $\phi \sim p(\phi)$ & $\phi \sim p(\phi)$ \\
        & $\v z \sim \mathcal{N}_p(\v 0, \m I_p)$\\
        $\v\theta|\phi \sim\mathcal{N}(\v 0, \m k_{\v x}(\phi))$ & $\v\theta = \v\mu_{\v\psi}(\v z, \phi)$\\
        $\v y \sim \mathcal{N}(\v\theta, \sigma)$ & $\v y \sim \mathcal{N}(\v\theta, \sigma)$ \\ \hline
    \end{tabular}
    \caption{True and PriorCVAE model comparison for a Gaussian Process regression example.}
    \label{tab:GPexample}
\end{table}
In this example, $\m k_{\v x}(\phi)$ is some kernel function evaluated at points $\v x$ for a given value of the length scale $\phi$. The VAE is trained using prior draws from the true model. In this instance, the model approximation provided by PriorCVAE replaces the GP prior with a deterministic reconstruction of the GP given $\v z$ and $\phi$,  
thus reducing the dimensionality of the inference task to be proportional to the dimensionality of $\v z$ 
rather than $\v \theta$. The latter of which scales with the number of observations in the case of a GP regression. 

 \section{Structured Independence via Generative Model Approximations}\label{section:methods}

 In this section, we introduce the \emph{\textbf{S}tructured \textbf{I}ndependence via \textbf{G}enerative \textbf{M}odel \textbf{A}pproximation} (SIGMA) prior. A SIGMA prior approximates the latent structure of a probabilistic model using a VAE with a generative model that satisfies desired conditional independence properties. We motivate the use of SIGMA priors FL scenarios where the prior structure of the desired model does not factorize over clients. We learn a SIGMA prior with the desired conditional independence and factorization properties, which replaces the original prior. The resulting approximate probabilistic model with the SIGMA prior is then amenable to use with existing FL algorithms, such as SFVI (Section~\ref{section:SFVI}). 

\subsection{SIGMA Prior}\label{section:sigma_prior}
The SIGMA prior is a novel approach to approximate the hierarchical Bayesian model described by equations \eqref{eq:model_class}-\eqref{eq:difficult_density}. It introduces $J+1$ latent variables~$\v z_G, \v z_{L}$, where $\v z_L = (\v z_{L_1}, \ldots, \v z_{L_j})$, and $J$ is the number of clients, to capture both the shared and client-specific variations in the local parameters $\v\theta_{j}$. The global latent variable $\v z_{G}$ induces dependence across clients, while the local latent variables $\v z_{L_j}$ drive the client-specific variations. We choose the hierarchical structure to enable the desired conditional independence properties across clients, facilitating the use of computationally efficient FL algorithms. The form of the generative model is

\begin{equation}\label{eq:SIGMAlvm}
\begin{split}
\v \phi &\sim p(\v \phi), \\
\v z_G &\sim \tp(\v z_G), \\
\v z_{L_j} &\sim \tp(\v z_{L_j}), \quad j \in \{ 1, \ldots, J \}, \\
\v \theta_{j} &\sim \mathcal{N}_{d_j}(\v\mu_{\v\psi_{L_j},\v\psi_{G}}(\v z_{L_j}, \v \phi), \gamma I_{d_j}), \quad j \in \{ 1, \ldots, J\},
\end{split}
\end{equation}
where $\v \phi\in\mathbb{R}^{d}$, $\v z_G\in\mathbb{R}^{p}$, $\v z_{L_j}\in\mathbb{R}^{p_{j}}, \v \theta_{j}\in\mathbb{R}^{d_j}$  for~$j\in \{1, \ldots, J \}$, and $\gamma > 0$. The $J$ functions $\v\mu_{\v\psi_{L_j},\v\psi_{G}}:\mathbb{R}^{p_j+p+d}\rightarrow \mathbb{R}^{d_j}$ are neural networks, with parameters $\v\psi_{L_j}$ and $\v\psi_{G}$ 
respectively, 
used to produce the decoder for the VAE.

\begin{figure}[ht!]
\centering
\includegraphics[width=0.95\textwidth]{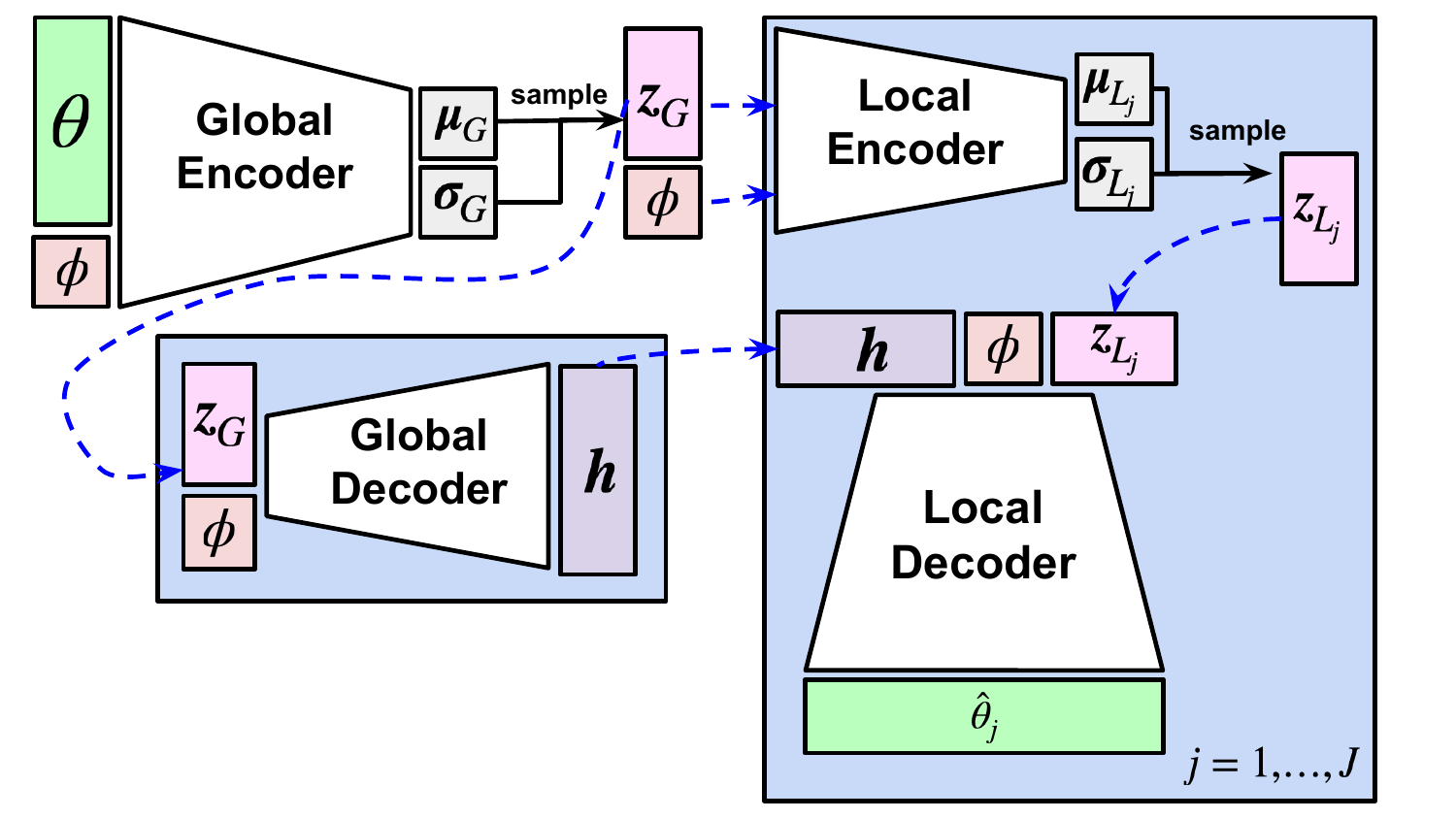}
\caption{Graphical description of the generative model fit to input prior draws 
$\v\theta$ to create the SIGMA prior. 
The output is $\hat{\v\theta}=(\hat{\v\theta}_1^\top, \ldots, \hat{\v\theta}_J^\top)^\top$, where each of the $J$ blocks of parameters $\hat{\v\theta}_{j}$ is conditionally independent given the global latent variable $\v z_G$ and the deterministic output $\v h$ of a neural network $f_{\v\psi_G}$, the global decoder,
that takes $\v z_G$ and $\phi$ as input. The SIGMA prior is trained as a hierarchical variational autoencoder with a specific structure, such that the local components of the encoders and decoders factorize over clients.}
\label{fig:SIGMA_structure}
\end{figure}

In this work, we employ a hierarchical structure for the decoders, where a \emph{global decoder} $\v g_{\v\psi_{G}}:\mathbb{R}^{p+d}\rightarrow \mathbb{R}^k$ is used as input to each \emph{local decoder} $\v f_{\v\psi_{L_j}}:\mathbb{R}^{p_j+d+k}\rightarrow \mathbb{R}^{d_j}$ with the following form,
\begin{equation}\label{eq:sigmadecoder}
\v\mu_{\v\psi_{L_j},\v\psi_{G}}(\v z_{L_j}, \v \phi) = \v f_{\v\psi_{L_j}}(\v z_{L_j}, \v \phi, \v g_{\v\psi_{G}}(\v z_{G}, \v \phi)).
\end{equation}
This hierarchical structure aids high-dimensional learning by allowing each local decoder to focus on learning the local deviations from a shared global representation captured by the global decoder $\v g_{\v\psi_{G}}$. The scale parameter $\gamma$ completes the overall decoder specification, denoted as $\mathcal{D}_{\v\psi} = (\v f_{\v\psi_{L_1}},\ldots, \v f_{\v\psi_{L_J}}, \v g_{\v\psi_{G}}, \log \gamma)$. Figure \ref{fig:SIGMA_structure} illustrates the generative model used to create the SIGMA prior, approximating the prior density \eqref{eq:difficult_density}. 

To learn the generative model, we approximate the posterior of the latent variables using a variational family $\tq_{\v\varphi}(\v z_L,\v z_{G} \vert \v \theta,\v \phi)$
with the following structure, 
\begin{equation}\label{eq:SIGMAvar}
    \begin{split}
        (\v z_G \vert \v \phi, \v \theta) & \sim \mathcal{N}_p(\v\mu_{\v \varphi_G}(\v\theta,\v\phi), \v\sigma_{\v \varphi_G}(\v\theta,\v\phi)) \\
        (\v z_{L_j} \vert \v z_G, \v \phi) & \sim \mathcal{N}_{p_j}(\v\mu_{\v \varphi_{L_j}}(\v z_G,\v\phi), \v\sigma_{\v \varphi_{L_j}}(\v z_G,\v\phi)), \quad j \in \{1,\ldots,J\},
    \end{split}
\end{equation}
which factorizes as $\tq_{\v\varphi}(\v z_L,\v z_{G} \vert \v \theta,\v \phi) = \tq_{\v\varphi_G}(\v z_G|\v \theta, \v\phi)\prod_{j=1}^J \tq_{\v \varphi_{L_j}}(\v z_{L_j}|\v z_G, \v\phi)$. We chose the factorization to mirror the conditional independence properties of the SIGMA prior, enabling efficient learning and inference. The \emph{global encoder} is parameterized by $\v\mu_{\v \varphi_G}:\mathbb{R}^{d_1 + \cdots + d_J + d} \rightarrow \mathbb{R}^p$ and $\v\sigma_{\v \varphi_G}:\mathbb{R}^{d_1 + \cdots + d_J + d} \rightarrow \mathbb{R}^p$, which are jointly output by a neural network. Similarly, the 
\emph{local encoders} are parameterized by $\v\mu_{\v \varphi_{L_j}}:\mathbb{R}^{p + d} \rightarrow \mathbb{R}^p$ and $\v\sigma_{\v \varphi_{L_j}}:\mathbb{R}^{p + d} \rightarrow \mathbb{R}^p$ for each $j\in \{1,\ldots,J\}$. The complete encoder structure is denoted as $\mathcal{E}_{\v\varphi} = (\v\mu_{\v \varphi_{L_1}},\ldots,\v\mu_{\v \varphi_{L_J}},\v\sigma_{\v \varphi_{L_1}},\ldots,\v\sigma_{\v \varphi_{L_J}},\v\mu_{\v \varphi_G},\v\sigma_{\v \varphi_G})$. We provide the architecture details of the encoders and decoders used in the numerical experiments in Section \ref{sec:experiment_details}.

\begin{table}
    \centering
    \begin{tabular}{lccc} \hline
         & Generative & Variational & Prior \\ 
         & model & posterior & approximation  \\\hline
         PriorVAE  & $\begin{aligned} \v z 
         &\sim \mathcal{N}_p(\v 0,I_p) \\ (\v\theta \vert \v z) &\sim \mathcal{N}_d(\v\mu_{\v\psi}(\v z),\gamma I_d)\end{aligned}$ & $\begin{aligned}(\v z &\vert \v\theta) \sim \\ &\mathcal{N}_p(\v\mu_{\v\varphi}(\v\theta),\v\sigma_{\v\varphi}
         (\v\theta))\end{aligned}$  & $\begin{aligned} \v z 
         &\sim \mathcal{N}_p(\v 0,I_p) \\ \v\theta &= \v\mu_{\v\psi}(\v z)\end{aligned}$ \\ \hline
         PriorCVAE   & $\begin{aligned} \v\phi 
         & \sim p(\v\phi) \\ \v z 
         & \sim \mathcal{N}_p(\v 0,I_p) \\ (\v\theta \vert \v z, \v\phi) & \sim \mathcal{N}_d(\v\mu_{\v\psi}(\v z, \v\phi),\gamma I_d)\end{aligned}$ & $\begin{aligned}(\v z &\vert \v\theta, \v\phi) \sim \\ &\mathcal{N}_p(\v\mu_{\v\varphi}(\v\theta,\v\phi),\v\sigma_{\v\varphi}(\v\theta,\v\phi))\end{aligned}$  & $\begin{aligned} \v\phi &\sim p(\v\phi) \\ \v z 
         &\sim \mathcal{N}_p(\v 0,I_p) \\ \v\theta &= \v\mu_{\v\psi}(\v z, \v\phi)\end{aligned}$ \\ \hline
         SIGMA Prior & Eq. \eqref{eq:SIGMAlvm}  & Eq. \eqref{eq:SIGMAvar}  & Eq. \eqref{eqn:SIGMA_model_approximation}  \\ 
         & & & \\ \hline
    \end{tabular}
    \caption{Comparison of Prior(C)VAE with SIGMA Prior: the columns show the generative model, the variational posterior approximation, and the approximation to the prior used for parameter inference.}
    \label{tab:priorcomparisons}
\end{table}

\subsection{Approximation of the prior}
We construct the SIGMA prior approximation from the estimated generative model and variational posterior of the VAE. In particular, we use
\begin{equation}\label{eqn:SIGMA_model_approximation}
\begin{split}
\v \phi &\sim p(\v \phi), \\
\v z_G &\sim \tp(\v z_G), \\
\v z_{L_j} & \sim \mathcal{N}_{p_j}(\v\mu_{\v \varphi_{L_j}}(\v z_G,\v\phi), \v\sigma_{\v \varphi_{L_j}}(\v z_G,\v\phi)), \quad j \in \{1,\ldots,J\}, \\
\v \theta_{j} &\sim \mathcal{N}_{d_j}(\v\mu_{\v\psi_{L_j},\v\psi_{G}}(\v z_{L_j}, \v z_{G}, \v \phi), \gamma I_{d_j}), \quad j \in \{ 1, \ldots, J\},
\end{split}
\end{equation}
which is identical to the generative model \eqref{eq:SIGMAlvm}, except that the distribution of each local latent variable $\v z_{L_j}$ is replaced by its variational approximation. Using the variational approximation of $\v z_{L_j}$ as the prior is a significant advantage of the SIGMA prior compared to methods like PriorCVAE, as it allows for incorporating the learned variational approximation into the approximate prior. The hierarchical structure of the SIGMA prior enables this replacement, as the variational approximations for the local latent variables are conditioned only on the global latent variable $\v z_G$ and the global parameters $\v \phi$, which are already part of the approximate prior specification.

The SIGMA prior offers significant benefits by incorporating learned variational approximations for the local latent variables, which represent the optimal density of the latent space for reconstructing draws of $\v \theta$. These local latent variables capture important information about the structure and dependencies within $\v \theta$, enabling the SIGMA prior to effectively model and generate realistic samples, setting it apart from approaches like PriorVAE and PriorCVAE.


Model \ref{pseudocode:model} shows pseudocode for a one-dimensional GP regression. Model \ref{pseudocode:SIGMA} shows pseudocode for an \emph{approximation} to Model \ref{pseudocode:model} using a SIGMA prior, where the encoders and decoders are trained using prior draws from the length scale parameter $\phi$ and the random variables $\v\theta$, denoted in Model \ref{pseudocode:model} as \code{phi} and \code{theta} respectively. The pseudocode closely resembles how models are defined in modern Python probabilistic programming languages such as Pyro \citep{bingham2019pyro}, Numpyro \citep{phan2019composable}, and PyMC \citep{abril2023pymc}. The function call \code{deterministic} is employed to define variables whose values are computed directly from other variables without any inherent randomness. The function call \code{sample} is used for drawing random variables from probability distributions, and the context \code{plate} represents a mechanism for indicating conditional independence across different subsets of random variables and data.
\begin{lstlisting}[language=Python, basicstyle=\ttfamily\small, caption=Pseudocode for a 1D GP regression model,label=pseudocode:model]
def model(y, x, k):
    # lengthscale prior
    phi = sample(Uniform(0.01, 1))
    # observation noise prior
    sigma = sample(HalfNormal(0, 1))
    # form covariance matrix
    cov = deterministic(k(x, x, phi))
    # GP prior
    theta = sample(MultivariateNormal(0, cov))
    # likelihood
    sample(Normal(theta, sigma), observed=y)
\end{lstlisting}

\begin{lstlisting}[language=Python, basicstyle=\ttfamily\small, caption=Pseudocode for the SIGMA prior approximation of a 1D GP regression model,label=pseudocode:SIGMA]
def approximate_model(ys, mus, encoders, decoders):
    # lengthscale prior
    phi = sample(Uniform(0.01, 1))
    # observation noise prior
    sigma = sample(HalfNormal(0, 1))
    # prior for global latents 
    z_G = sample(Normal(0, 1))
    # variables inside plate are conditionally independent across clients
    with plate("clients", len(mus)):
        # prior for local latents
        mu[j], logvar[j] = deterministic(encoders[j](z_G, phi))
        z_L[J] = sample(Normal(mu[j], exp(0.5*logvar[j])))
        # estimate likelihood mean 
        theta[j] = deterministic(decoders[j](z_G, z_L[j], phi))
        # likelihood
        sample(Normal(theta[j], sigma), observed=y[j])
        
\end{lstlisting}

The SIGMA prior approximation enables the inference of the latent variables $\v z_G, \v z_{L_1}, \ldots, \v z_{L_J}$ instead of the original local parameters $\v \theta_{1}, \ldots, \v \theta_{J}$. Notably, the SIGMA prior ensures the required conditional independence: $\hat{\v \theta}_{j} \perp\!\!\!\perp \hat{\v \theta}_{k} \mid \v z_G, \v\phi$ for all pairs of clients $j \neq k$. This property is useful in FL settings, as it allows each client to perform local updates independently without the need for sharing raw data, thereby facilitating efficient and privacy-preserving learning. Furthermore, this property enables the use of existing FL algorithms such as SFVI \citep{hassan2023federated} or FedSOUL \citep{kotelevskii2022fedpop}, which rely on the assumption of conditional independence among local parameters given the global parameters.

\section{Numerical Examples}\label{section:numerical_examples}
This section presents two numerical experiments that showcase the effectiveness of the SIGMA prior in handling highly dependent priors with increasing dimensionality and numbers of clients. These experiments demonstrate the scalability and adaptability of our approach in various settings, ranging from simple one-dimensional examples to complex real-world spatial structures.

The first experiment establishes a baseline for the SIGMA prior's performance using a one-dimensional GP prior with $p=100$ dimensions and $J=3$ clients. We provide qualitative assessments through visualizations, including the empirical covariance of the SIGMA prior compared to the covariance of the GP prior, reconstructed draws from the SIGMA prior, and comparison of estimates when using the SIGMA prior to construct an approximate model compared to the true underlying model that contains the GP prior.
The second experiment demonstrates the SIGMA prior's ability to handle complex, real-world spatial structures by considering a prior structure with dimension $p=2227$ and $J=8$ clients, motivated by the Australian geography. This experiment showcases the scalability of our approach to more significant and realistic problems.

Throughout these experiments, we employ stochastic variational inference implemented in the Pyro library \citep{bingham2019pyro} to train the VAE, with encoders and decoders constructed using PyTorch \citep{paszke2019pytorch}. For posterior estimation in the approximate model, we use the \emph{No-U-Turn sampler} \citep[NUTS,][]{hoffman2014no} algorithm in the first example, and mean-field variational inference in the second example. Both algorithms are implemented in the BlackJAX \citep{cabezas2024blackjax} library.

To evaluate the performance of the SIGMA prior, we use a combination of qualitative and quantitative metrics tailored to each experiment. Experiment 1 focuses on visual assessments.
In Experiment 2, we compare inference estimates from the approximate SIGMA prior model with MCMC estimates of the true model. Additionally, we propose a novel correction by viewing the approximate model created with the SIGMA prior as an auxiliary-variable model, emphasizing how the SIGMA prior enables similar parameter estimates using an FL algorithm compared to MCMC and VI results on the actual model.

The following subsections (Sections \ref{sec:experiment_1} and \ref{sec:experiment_3}) present each experiment in detail, discussing the specific settings, results, and insights gained. Appendix \ref{sec:experiment_details} describes the architecture details of the VAE common to the two experiments.

\subsection{One-dimensional Gaussian Process Regression}\label{sec:experiment_1}

In our first example, we demonstrate the effectiveness of the SIGMA prior in approximating a one-dimensional GP regression. We generate training data by sampling from a GP with a radial basis function (RBF) kernel at 100 equidistant points $\v x_{\text{train}}$ in the interval $[0, 1]$. The data is divided among three clients based on the input values: client 1 receives data associated with $\v x \in [0, 0.34)$, client 2 receives data associated with $\v x \in [0.34, 0.67)$, and client 3 receives data associated with $\v x \in [0.67, 1.00]$.

We generate $10,000$ draws from a GP prior with an RBF kernel with the lengthscale parameter varying uniformly between $0.2$ and $1$ and with variance fixed at $1$. We train the SIGMA prior architecture for $20,000$ iterations using the Adam optimizer \citep{kingma2014adam} with a learning rate of $10^{-3}$ in Pyro \citep{bingham2019pyro}. The batch size in each iteration is $32$, and the dimension of the hidden layers used in the neural net architectures is $16$. The dimension of the global latent space $\v z_G$ and each of the three local latent spaces $\v z_{L_1}, \v z_{L_2}, \v z_{L_3}$ were chosen to be $5$ respectively, and the standard deviation of the decoder is set to $\gamma=0.5$. 

\begin{figure}[h]
    \centering
\includegraphics[width=0.96\textwidth]{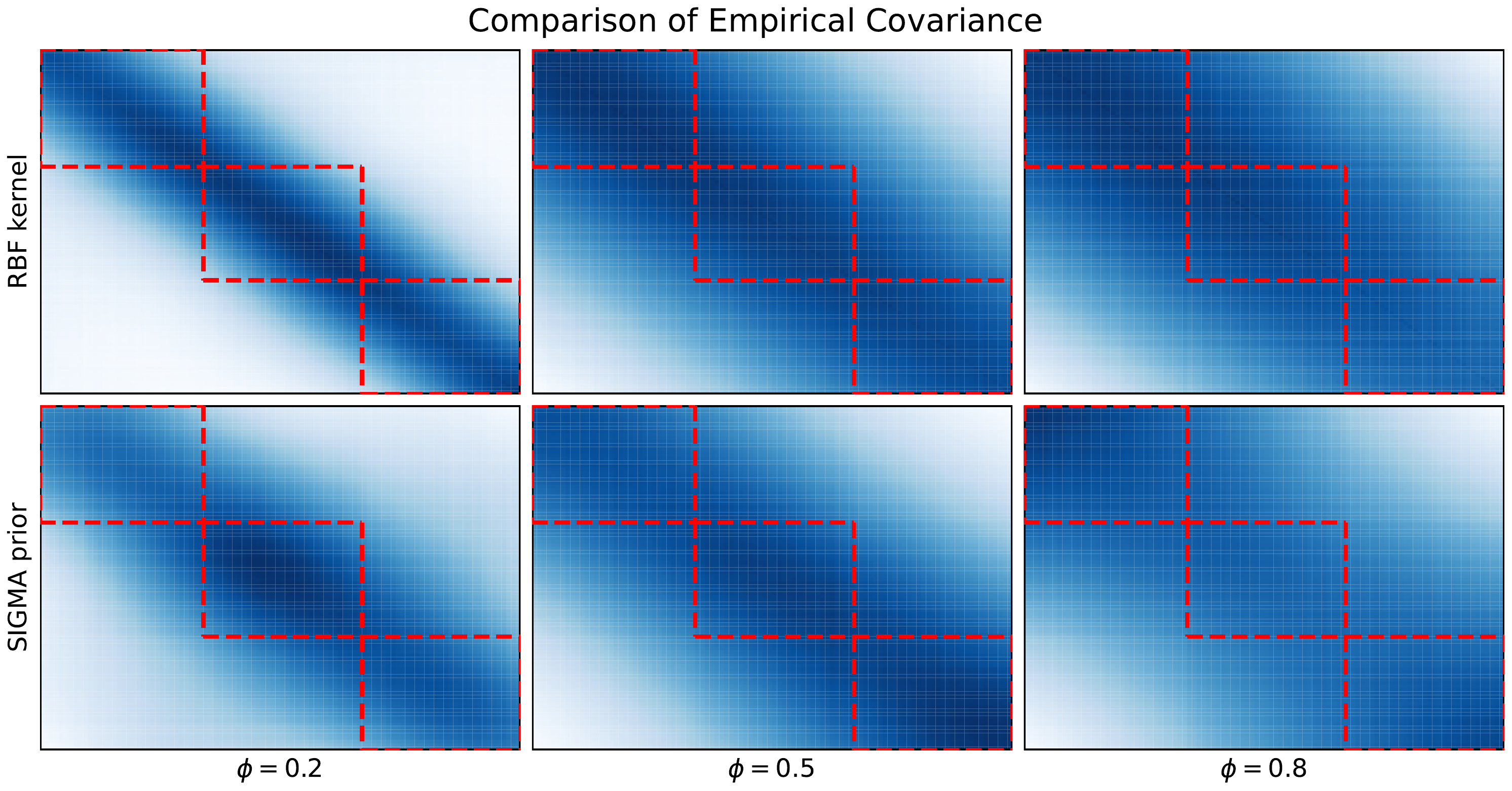}
    \caption{Comparison of the empirical covariance of the draws from the RBF kernel (top row) and the learned SIGMA prior (bottom row) for $\phi=(0.2, 0.5, 0.8)$. The SIGMA prior closely approximates the true covariance structure, capturing client dependencies.}
    \label{fig:1d_empirical_covariance}
\end{figure}

We train the SIGMA prior and compare its empirical covariance with the actual RBF kernel. Figure \ref{fig:1d_empirical_covariance} visualizes this comparison, demonstrating that the learned SIGMA prior closely approximates the actual covariance structure for varying values of the length scale parameter.

\begin{figure}[h]
\centering
\includegraphics[width=0.8\textwidth]{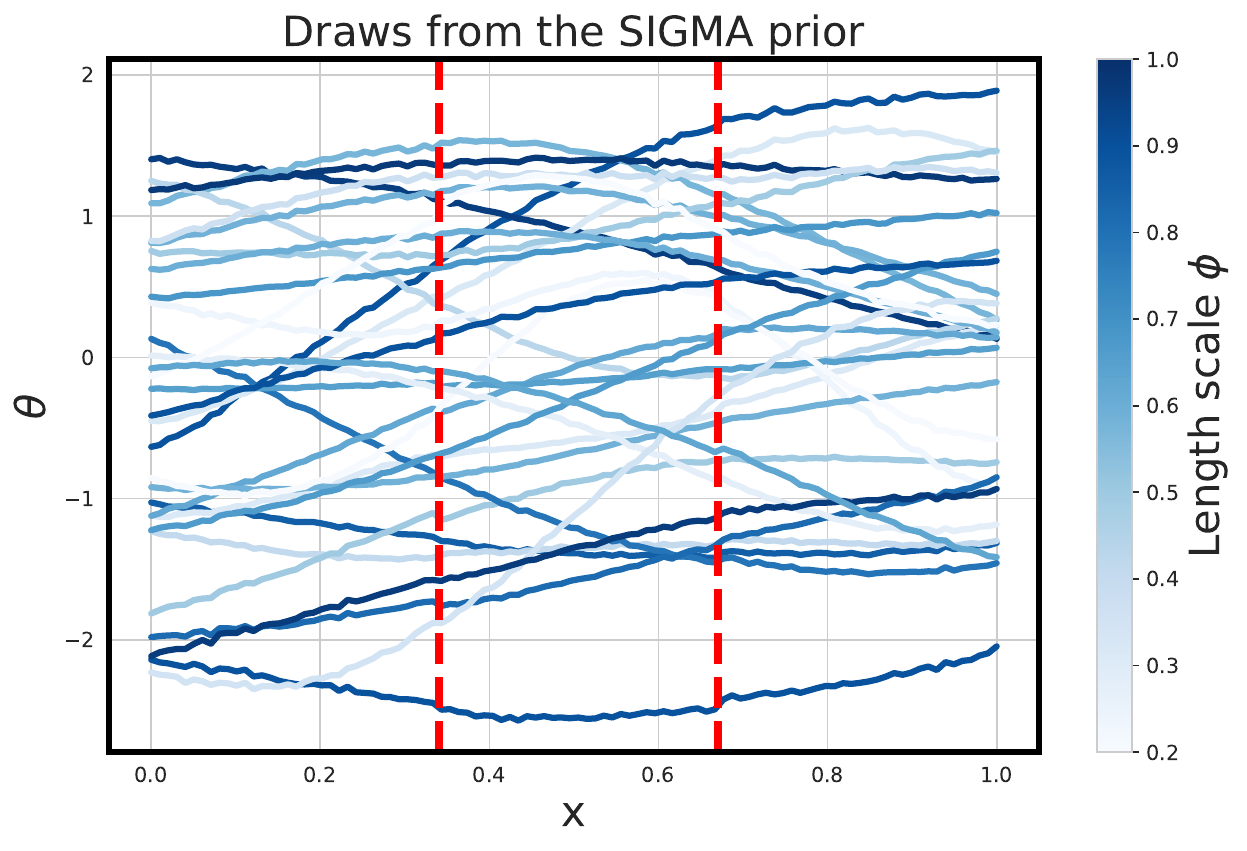}
\caption{$25$ draws from the fitted SIGMA prior. The red-vertical-dashed lines denote boundaries between clients, and the colors of each line denote the length scale of the Gaussian process kernel that the particular line is approximating.}
\label{fig:1d_prior_draws}
\end{figure}

To further illustrate the SIGMA prior's ability to capture dependencies across clients, we generate $25$ prior draws from the fitted SIGMA prior (Figure \ref{fig:1d_prior_draws}). The red-vertical-dashed lines denote the boundaries between clients, and the smooth transitions across these boundaries demonstrate that the SIGMA prior has learned the underlying correlations in the data.

\begin{figure}[h]
\centering
\includegraphics[width=0.8\textwidth]{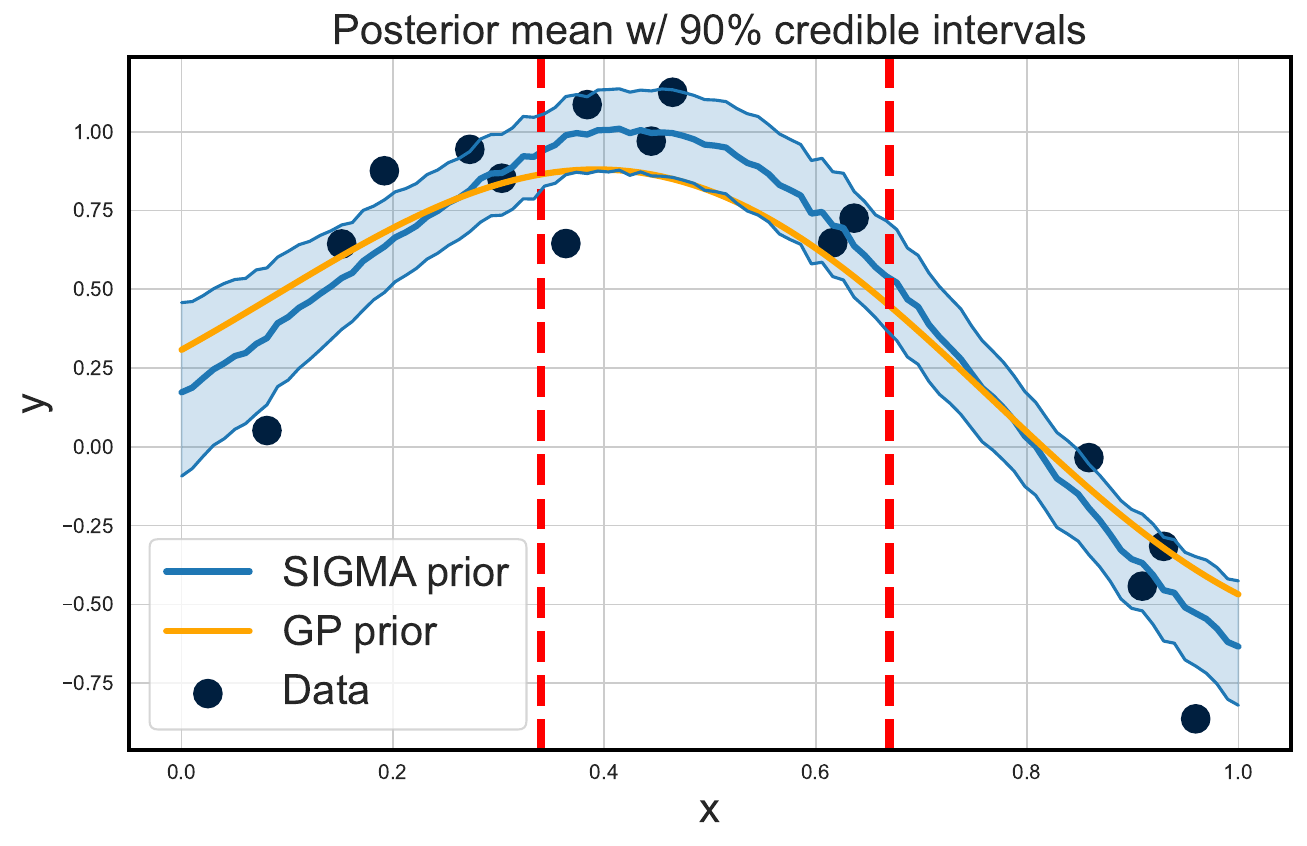}
\caption{Mean estimate (solid blue line) and 90\% credible interval (shaded region) for the mean function using the SIGMA approximation. The posterior was estimated using the NUTS algorithm. The red dashed lines separate the different clients.}
\label{fig:1d_GP_posterior}
\end{figure}

We now use the SIGMA approximation in an inferential model. We generate 15 random locations $\v x$ in the interval $[0, 1]$ and observations $\v y = \sin(\v x) + 0.2 \cdot {\v\epsilon},$ where $\v \epsilon\sim\mathcal{N}(0, 1)$. Figure \ref{fig:1d_GP_posterior} shows the mean estimate (solid blue line) and 90\% credible interval (shaded region) for the mean function using the SIGMA approximation, with the posterior estimated using the NUTS algorithm.

This example showcases the effectiveness of the SIGMA prior in approximating a one-dimensional GP regression while capturing dependencies across clients in a FL setting. The learned SIGMA prior closely approximates the covariance structure, generates prior draws that smoothly transition across client boundaries, and provides a reasonable estimate of the posterior mean.

\subsection{Fitting Bayesian spatial models over the Australian geography}\label{sec:experiment_3}
In this experiment, we aim to demonstrate the effectiveness of the SIGMA prior in approximating the spatial structure of the Australian geography, using a proper conditional autoregressive (PCAR) prior (see Appendix \ref{PCAR}) to model dependencies at the Statistical Area Level 2 (SA2). This experiment is motivated by the Australian Cancer Atlas \citep[ACA,][]{duncan2019development}, which produces estimates of cancer diagnoses, excess deaths, and risk factors across the Australian geography at the SA2 level. The ACA comprises over one hundred individual spatial and spatio-temporal models, each employing a form of a CAR prior to smooth estimates and to account for spatial dependence in the data. Integrating new data sources into the ACA poses challenges, as it requires establishing new data agreements across the Australian geography. The SIGMA prior offers a potential solution by enabling the approximation of complex spatial structures while facilitating efficient and privacy-preserving learning in a federated setting.

In this experiment, we consider $J=8$ clients representing Australia's eight states and territories. The parameter vector $\v\theta\in\mathbb{R}^{2221}$ is assigned a PCAR prior distribution, $\v\theta|\phi \sim \mathcal{N}(\v 0, \m\Sigma^{-1})$, where $\m\Sigma\in\mathbb{R}^{2221\times 2221}$ is the precision matrix defined as $\m\Sigma = \m D - \phi \m W$, where $\m W$ is the adjacency matrix, with $\m W_{ij}=1$ if observations $i$ and $j$ are neighbors and $\m W_{ij}=0$ otherwise. The diagonal matrix $\m D$ captures the number of neighbors for each observation, with $\m D_{ii}$ representing the number of neighbors for observation $i$. The unknown parameter $\phi\in (0, 1)$ is interpreted as a measure of the spatial smoothing between neighbors, with higher values indicating stronger spatial dependence. 

We generate $10,000$ draws from the PCAR prior over the Australian geography, with $\phi$ varying uniformly between $0$ and $1$. We train the SIGMA prior architecture for $400,000$ iterations using the Adam optimizer \citep{kingma2014adam} with a learning rate of $1e^{-4}$ in Pyro \citep{bingham2019pyro}. The batch size in each iteration is $128$, and the dimension of the hidden layers used in the neural net architectures is $2024$ for the decoders and the global encoder and $1048$ for the local encoders. The dimension of the global latent space $\v z_G$ is $100$ and the dimension for each of the local latent spaces is $50$. The standard deviation of the decoder is set to $\gamma=0.03$. 

Due to the SIGMA prior being a model approximation and assessing the performance of an approximate algorithm in VI, we chose to use a synthetic dataset for this example where the response is a draw from the PCAR prior over the Australian geography. The synthetic dataset was generated by sampling from the PCAR prior. The precision matrix $\m\Sigma$ was constructed using the adjacency matrix $\m W$ derived from the SA2-level spatial structure of Australia, ensuring that the synthetic data accurately reflects the complex spatial dependencies present in the real-world geography. By assessing the approximations of the SIGMA prior and VI against this synthetic dataset, we can effectively evaluate their performance in capturing the true data-generating process without the confounding effects of model misspecification or data quality issues.

We employ \citep[NUTS,][]{hoffman2014no} and mean-field variational inference (MFVI) algorithms, both implemented in BlackJAX \citep{cabezas2024blackjax}, to estimate the posterior distribution of the model parameters. For NUTS, we run the sampler for $5,000$ iterations using the default hyperparameters and summarize the results using the last $2,500$ draws of the chain, discarding the first $2,500$ iterations as warmup. For MFVI, we run the algorithm for $10,000$ iterations, utilizing the Adam optimizer \citep{kingma2014adam} with a learning rate of $1e^{-3}$. We clip the gradients if their norm exceeds $1e^{-3}$ to ensure stability.

\subsubsection{True underlying model}\label{sec:aca_true_model}
To assess the effectiveness of the SIGMA prior in approximating complex spatial dependencies, we define a true underlying model that serves as the target for our approximation. The true model for this example is 
\begin{align}\label{eqn:aca_true_model}
\text{logit } \phi &\sim \mathcal{N}(0, 2), \\
\v\theta | \phi &\sim \mathcal{N}\big (\v 0, (\m D - \phi \m W)^{-1}\big ), \\
\v y | \v \theta &\sim \mathcal{N}(\v\theta, 0.5).
\end{align}
The spatial smoothing parameter $\phi$ is transformed using the logit function to ensure it remains within the valid range of (0, 1). The prior distribution for $\text{logit } \phi$ is a Gaussian distribution with mean 0 and variance 2, allowing for a wide range of possible values for $\phi$.

The spatial random effects $\v\theta$ are assigned a PCAR prior, conditional on $\phi$. The precision matrix $(\m D - \phi \m W)^{-1}$ captures the spatial dependence structure, where $\m D$ and $\m W$ are the diagonal matrix of neighbor counts and the adjacency matrix, respectively, as described in the previous section. The observed data $\v y$ are modeled using a normal likelihood function, conditional on the spatial random effects $\v\theta$. The standard deviation of the likelihood function is set to $0.5$, representing the assumed measurement error or uncertainty in the observations.

\subsubsection{Deterministic SIGMA prior model}\label{sec:aca_ex_sigma_prior}
After training the SIGMA prior on prior draws from the true PCAR model across the Australian geography, we obtain the following approximation, which we refer to as the \emph{deterministic SIGMA prior model}, 
\begin{align}\label{eqn:aca_deterministic_model}
\text{logit }\phi &\sim \mathcal{N}(0, 2), \\
\v z_G &\sim \mathcal{N}(0, 1), \\
\v z_{L_j} | \v z_G, \phi &\sim \mathcal{N}(\v\mu_{\v \varphi_{L_j}}(\v z_G, \phi), \v\sigma_{\varphi_{L_j}}(\v z_G, \phi)), \quad j = {1, \ldots, J}, \\
\hat{\v\theta}_j &= \v\mu_{\v\psi_{L_j}, \v\psi_G}(\v z_{L_j}, \v z_G, \phi), \quad j = {1, \ldots, J}, \\
\v y_{j} | \hat{\v\theta}_{j} &\sim \mathcal{N}(\hat{\v\theta}_j, 0.5). 
\end{align}
From here on, we refer to this form of the approximate model as the ``\emph{deterministic SIGMA prior model}'', as reconstructed estimates of $\v\theta$ are deterministic given the other unknown parameters. In this approximation, the global latent variable $\v z_G$ captures the shared information across all regions, while the local latent variables $\v z_{L_j}$ capture the client-specific information. The deterministic functions $\v\mu_{\v \varphi_{L_j}}$ and $\v\sigma_{\varphi_{L_j}}$ represent the encoder networks that map the global latent variable and the spatial smoothing parameter to the mean and standard deviation of the local latent variables, respectively. The decoder network $\v\mu_{\v\psi_{L_j}, \v\psi_G}$ maps the local and global latent variables, along with the spatial smoothing parameter, to the reconstructed estimates of the spatial random effects $\hat{\v\theta}_j$.

\begin{figure}[h]
    \centering
\includegraphics[width=0.95\textwidth]{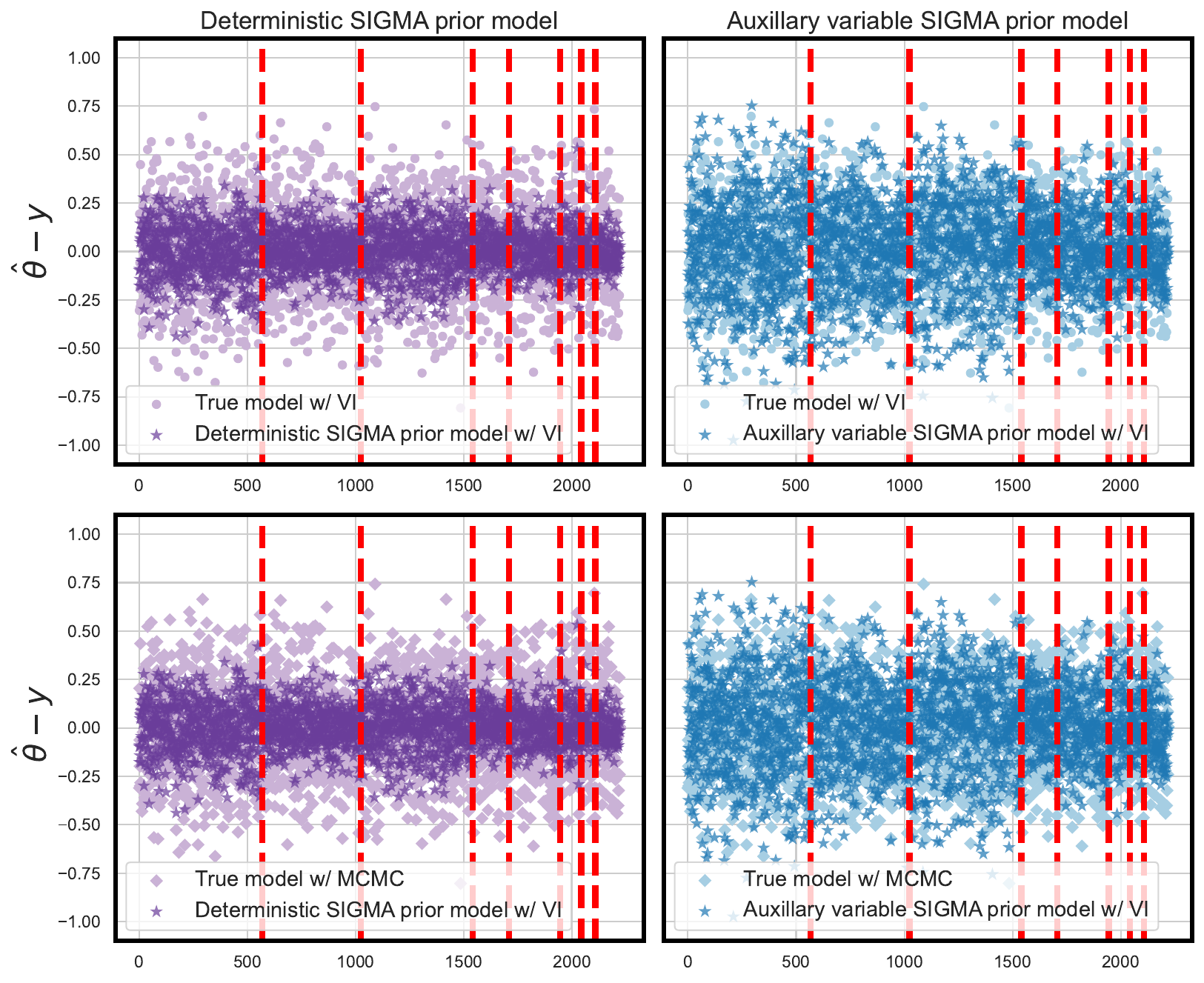}
    \hfill
    \caption{Plots compare the difference between the observed data and mean estimates for $\v\theta$, under each combination of model and algorithm. The top row shows the true model using VI, while the bottom row shows the true model using MCMC. The left-hand column shows the deterministic SIGMA prior model, and the right-hand column shows the auxiliary variable SIGMA prior model.}
    \label{fig:ACA_data}
\end{figure}

The plots on the left-hand side of Figure \ref{fig:ACA_data} show the difference between $\hat{\v\theta}$ and $\v y$ for the deterministic SIGMA prior model fitted using MFVI. Compared to the same estimate from the true model using MFVI (top) and the true model using NUTS (bottom), we observe that the estimates of $\v\theta$ from the deterministic SIGMA prior model overfit to the data. This overfitting behavior can be attributed to lower variance in prior draws of $\v\theta$ in the deterministic SIGMA prior model relative to the true model. Figure \ref{fig:ACA_MCMC} (left) further highlights this issue by showing that the difference in mean estimates of $\v\theta$ between the deterministic SIGMA prior model and MCMC estimates from the true model has much greater variance compared to the difference between the MCMC and VI estimates under the true model.

\subsubsection{Auxiliary variable SIGMA prior model}\label{sec:aca_ex_extended_inference}

Here, we propose an extension to the inferential model one would fit when using a model approximation with a SIGMA prior. The motivation of the extension is to reduce the tendencies of the approximate model to overfit to the observed data and to correct the narrow credible intervals. We note that these overfitting properties were observed and left as outstanding problems in both the PriorVAE and PriorCVAE papers, with overfitting appearing to a lesser extent in the PriorCVAE paper.

Our proposed extension is motivated by first considering the same generative scenario as PriorVAE, where we have random variables $\v\theta$ that we want to approximate with an alternative generative model. In the works of PriorVAE, PriorCVAE, and this paper, at inference, we have treated $\v\theta$ as deterministic given the added latent variables. Alternatively, we can consider treating $\v\theta$ as a random variable, with mean given by the decoder and variance $\v\tau^2$. From this lens, the added latent variable can be viewed as an auxiliary variable, and we would end up conducting inference over both $\v\theta$ and the latent variables. If we choose the generative model
\begin{align*}
\v z &\sim \mathcal{N}_P(0, 1), \\
\v\theta | \v z &\sim \mathcal{N}_D \big (\v\mu_{\v\theta} + \m W_{\v\theta}^\top \v z, \v\tau^2\big ),
\end{align*}
with unknown parameters $\v\mu_{\v\theta}, \m W_{\v \theta}$, and $\v\tau$, we are fitting probabilistic PCA \citep[PPCA,][]{tipping1999probabilistic} to prior draws from $p(\v\theta)$, where the maximum likelihood estimate for $\v\tau^2$ is the average of the $D-P$ smallest eigenvalues of the empirical covariance matrix of draws from $p(\v\theta)$. If viewing from this perspective, then fitting a PPCA model to prior draws of $\v\theta$ and, at inference, jointly sampling from $p(\v\theta, \v z) = p(\v\theta)p(\v\theta | \v z)$ can be viewed as an approximate auxiliary variable method. If the dimension of $\v z$ is chosen such that $P=D$, then $\v\tau^2 \rightarrow 0$, and we essentially fit a deterministic location-scale transform to draws of $\v\theta$.

As the decoder we use is not an affine transformation as in PPCA but a neural network, we do not get analytical estimates for the variance $\v\tau^2$. However, we offer a post-hoc correction. Under the true model, we can sample from the prior distributions to generate draws of $\v\theta$, which we denote as $\tilde{\v\theta}_{\text{True}}$. We do the same for the deterministic SIGMA prior model and call these draws $\tilde{\v\theta}_{\text{SIGMA}}$. We assume that $\v\theta$ is Gaussian distributed, centred on the decoder estimates, with unknown variance $\v\tau^2$. We attempt to match the amount of prior variance using the SIGMA prior and the true model. Given our formulation, the variance is additive, i.e., $\text{Var}(\tilde{\v\theta}_{\text{True}})=\text{Var}(\tilde{\v\theta}_{\text{SIGMA}}) + \v\tau^2$. This expression enables us to estimate $\v\tau^2$ after fitting the VAE but before inference by taking prior draws from the two models and calculating $\v\tau^2 = \text{Var}(\tilde{\v\theta}_{\text{True}}) - \text{Var}(\tilde{\v\theta}_{\text{SIGMA}})$. As the estimator for $\v\tau^2$ might not be positive, we clip the estimates to be between the range $0.01$ and $100$. For this example, $99.8\%$ of the elements of $\v\tau^2$ did not require clipping.

We call this model the \emph{auxiliary variable SIGMA prior model} and, for this example, define it as
\begin{align*}
\text{logit } \phi &\sim \mathcal{N}(0, 2), \\
\v z_G &\sim \mathcal{N}(0, 1), \\
\v z_{L_j}|\v z_G, \phi &\sim \mathcal{N}(\v\mu_{\v \varphi_{L_j}}(\v z_G, \phi), \v\sigma_{\varphi_{L_j}}(\v z_G, \phi)), \quad j \in \{1, \ldots, J\},\\
\hat{\v\theta} &= \v\mu_{\psi_{L_j}, \psi_G}(\v z_G, \v z_{L_j}, \phi), \quad j \in \{1, \ldots, J\}, \\
\v\theta|\v z_G, \v z_{L_j}, \phi &\sim\mathcal{N}(\hat{\v \theta}, \v\tau), \quad j \in \{1, \ldots, J\},\\
\v y_j | \v \theta &\sim \mathcal{N}(\v\theta, 0.5), \quad j \in \{1, \ldots, J\}.
\end{align*}

The addition of the conditional density of $\v\theta|\v z_G, \v z_{L_j}, \phi$ does not introduce any additional dependencies between clients, as all the dependencies are already captured by estimating $\hat{\v\theta}$ through the learned SIGMA prior. Thus, including this density is straightforward from an FL perspective. While treating $\v\theta$ as a random variable increases the dimensionality of the model, the computational cost remains manageable when using a variational inference algorithm with a mean-field assumption for the variational family of $\v\theta$.

\begin{figure}[h]
    \centering
\includegraphics[width=0.95\textwidth]{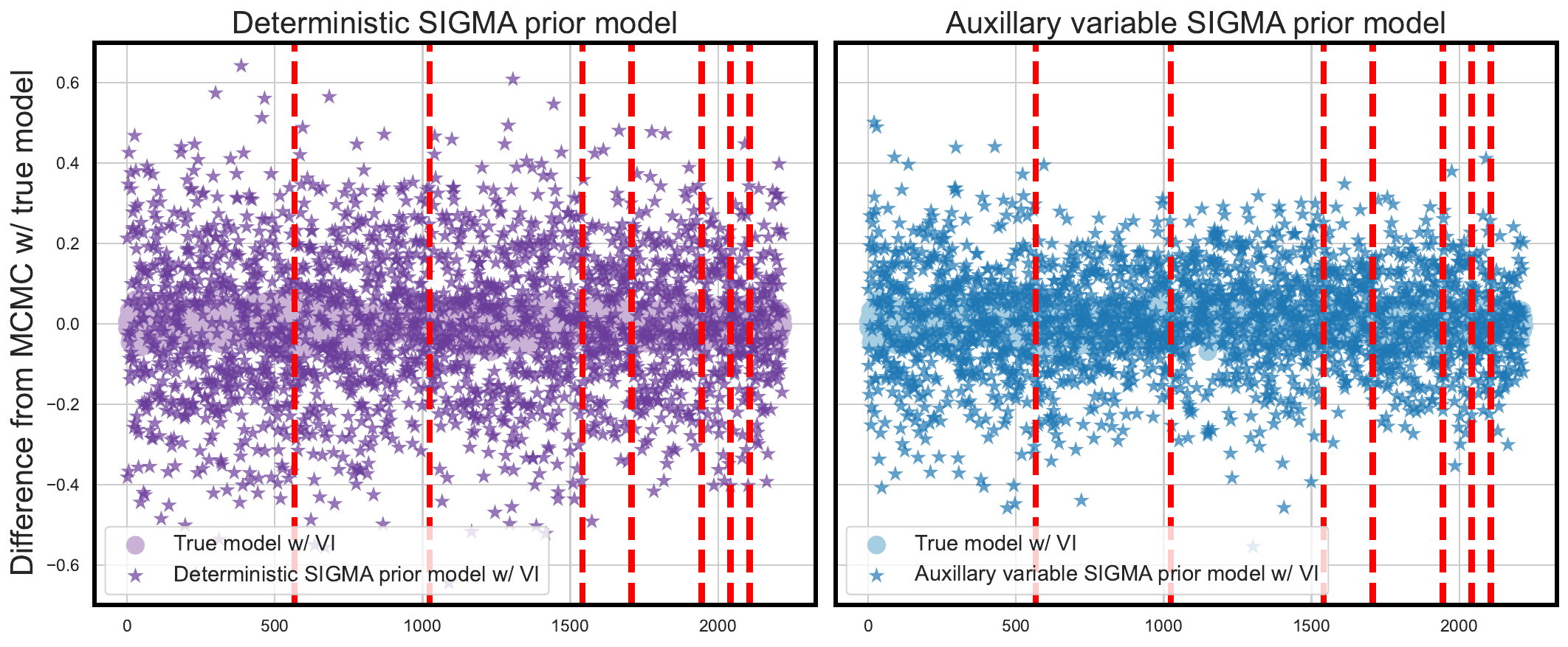}
    \hfill
    \caption{Comparison of the mean estimates of $\v\theta$ obtained from different models and inference methods. The vertical axis represents the difference between the mean estimates and the MCMC estimates of $\v\theta$ from the true model. Both plots include the difference between the mean estimates of $\v\theta$ using MFVI with the true model as a baseline. The left plot displays the difference between the mean estimates of $\v\theta$ using the deterministic SIGMA prior model and the MCMC estimates from the true model. The right plot shows the difference between the mean estimates of $\v\theta$ using the auxiliary variable SIGMA prior model and the MCMC estimates from the true model.}
    \label{fig:ACA_MCMC}
\end{figure}

Figure \ref{fig:ACA_data} (right-hand side) illustrates the difference between the mean estimate of $\v\theta$ and the observed data $\v y$ for the auxiliary SIGMA prior model fitted using mean-field variational inference (MFVI). Comparing these estimates to those obtained from the true model using MFVI (top) and NUTS (bottom) reveals that the auxiliary variable SIGMA prior model achieves similar levels of smoothing with respect to the data. Notably, the auxiliary variable approach significantly reduces the overfitting observed in the estimates from the deterministic SIGMA prior model. Moreover, Figure \ref{fig:ACA_MCMC} (right-hand side) demonstrates that the difference between the mean estimates of $\v\theta$ under the auxiliary variable SIGMA prior and the MCMC estimates under the true model is substantially smaller than the difference between the mean estimates of $\v\theta$ under the deterministic SIGMA prior and the MCMC estimates. This finding suggests that the auxiliary variable SIGMA prior model approximates the true posterior distribution more closely than the deterministic SIGMA prior model.

\section{Conclusion}\label{section:discussion}

In this paper, we introduced the 
\emph{\textbf{S}tructured \textbf{I}ndependence via deep \textbf{G}enerative \textbf{M}odel \textbf{A}pproxi\-mation}
(SIGMA) prior, a novel approach to approximating the latent structure of probabilistic models while enforcing desired conditional independence properties. The SIGMA prior utilizes a hierarchical variational autoencoder (VAE) to capture dependencies in the original prior specification and induce conditional independence among subsets of variables. The key components of the hierarchical VAE include a global latent variable that captures shared information across all clients, local latent variables that capture client-specific information, and encoder and decoder networks that map between the latent space and the original parameter space. By carefully designing the architecture and training objective of the VAE, the SIGMA prior enables the approximation of complex prior distributions while maintaining the desired conditional independence structure.

The development of the SIGMA prior is motivated by the challenges of horizontal FL, where data is distributed across multiple clients, and direct access to data from other clients is prohibited. Existing FL algorithms, such as SFVI \citep{hassan2023federated}, assume independence between clients, limiting their applicability to scenarios where dependencies are present. The SIGMA prior addresses this limitation by approximating the prior distribution over client-specific parameters and inducing conditional independence between clients, enabling the use of existing FL algorithms while capturing dependencies between clients. This advancement is significant, as it allows for accurate inference and prediction in real-world applications where dependencies between clients are expected.
We evaluated the performance of the SIGMA prior on a synthetic and real-world application, including a one-dimensional GP prior and a CAR prior for modeling spatial dependencies in cancer estimates across the Australian geography. Our experiments showed that the SIGMA prior effectively captures the dependencies in the original prior distributions and provides a close approximation to estimates from the actual model in the non-FL setting. These results highlight the potential of the SIGMA prior as a general-purpose tool for enabling FL with dependent data. The successful application of the SIGMA prior to the CAR prior across the Australian geography demonstrates the real-world impact of our approach. Spatial dependencies are prevalent in many fields, such as epidemiology, environmental sciences, and social sciences, where data is often collected at different spatial resolutions or administrative units. The SIGMA prior provides a principled way to model these dependencies in an FL setting, enabling collaboration between organizations or data custodians without compromising data privacy. 

Despite the promising results, there are several limitations and avenues for future research. One avenue for potential extensions is to explore alternative generative models for the SIGMA prior, such as normalizing flows \citep{rezende2015variational}, generative adversarial networks \citep{goodfellow2014generative}, or denoising diffusion models \citep{song2019generative, ho2020denoising, kingma2021variational}. These models have shown promising results in capturing complex distributions and could improve the expressiveness and flexibility of the SIGMA prior. Additionally, the auxiliary variable SIGMA prior model introduced in this work offers a promising direction for future research. The auxiliary variable approach can reduce overfitting and improve the calibration of uncertainty estimates by treating the local parameters as random variables and introducing an additional conditional distribution. Further investigation into the theoretical properties and practical implications of the auxiliary variable SIGMA prior model could lead to more robust and reliable model approximations.

In conclusion, the SIGMA prior is a powerful and flexible tool for enabling FL with dependent data. By approximating the prior distribution and inducing conditional independence, the SIGMA prior allows using existing FL algorithms while capturing dependencies between clients. Our work opens up new possibilities for collaborative learning and privacy-preserving inference in various applications, from spatial modeling in healthcare and beyond. As FL continues to evolve, ideas surrounding the SIGMA prior could play an essential role in developing more expressive and accurate models that can leverage the full potential of decentralized data.
\bibliographystyle{apalike}
\bibliography{references}

\appendix
\section{Architecture details for the SIGMA prior}\label{sec:experiment_details}

The global encoder generates the variational parameters $\v\eta_G:=(\v\mu_G^{\top}, \v\tau_G^{\top})^{\top}$ for the posterior approximation of $\v z_G$ by taking input $\v g^G = (\v y^{\top}, \v\phi^{\top})^{\top}$ and applying the following functions:
\begin{align*}
\v f^G &= \m W_2^G (\sigma(\m W_1^G \v g^G + \v b_1^G)) + \v b_2^G, \\
\v\mu_G &= \m W_{\v\mu}^G\v f^G + \v b_{\v\mu}^G, \\
\log\v\tau_G^2 &= \m W_{\v\tau}^G \v f^G + \v b_{\v\tau}^G,
\end{align*}
where $\m W_1^G \in\mathbb{R}^{h_G\times (n+1)}, \v b_1^G \in\mathbb{R}^{h_G}, \m W_2^G \in\mathbb{R}^{h_G \times h_G}$, $\v b_2^G \in\mathbb{R}^{h_G}$, $\m W_{\v\mu}^G\in\mathbb{R}^{n_G\times h_G},
\v b_{\v\mu}^G \in\mathbb{R}^{n_G},
\m W_{\v\tau}^G \in\mathbb{R}^{n_G \times h_G},  \v b_{\v\tau}^G\in\mathbb{R}^{n_G}$, and $\sigma(\cdot)$ represents the ReLU activation function \citep{nair2010rectified} applied elementwise.
For each of the $J$ clients, the local encoder generates the variational parameters $\v\eta_{L_j}:=(\v\mu_{L_j}^{\top}, \v\tau_{L_j}^{\top})^{\top}$ for the posterior approximation of $\v z_{L_j}$ using:
\begin{align*}
\v g^{L_j} &= (\v z_G^{\top}, \v\phi^{\top})^{\top}, \\
\v f^{L_j} &= \m W_2^{L_j} (\sigma(\m W_1^{L_j} \v g^{L_j} + \v b_1^{L_j})) + \v b_2^{L_j},\quad j \in {1, \ldots, J},\\
\v\mu_{L_j} &= \m W_{\v\mu}^{L_j}\v f^{L_j} + \v b_{\v\mu}^{L_j}, \quad j \in {1, \ldots, J},\\
\log \v\tau_{L_j}^2 &= \m W_{\v\tau}^{L_j}\v f^{L_j} + \v b_{\v\tau}^{L_j}, \quad j \in {1, \ldots, J},
\end{align*}
where $\m W_1^{L_j} \in\mathbb{R}^{h_{L}\times (n_G + 1)}, \v b_1^{L_j} \in\mathbb{R}^{h_{L}}, \m W_2^{L_j} \in\mathbb{R}^{h_L \times h_L}$, $\v b_2^{L_j} \in\mathbb{R}^{h_L}$, $\m W_{\v\mu}^{L_j}\in\mathbb{R}^{n_{L_j}\times h_{L}}, \v b_{\v\mu}^{L_j} \in\mathbb{R}^{n_{L_j}},\m W_{\v\tau}^{L_j}\in\mathbb{R}^{n_{L_j}\times h_{L}},  \v b_{\v\tau}^{L_j} \in\mathbb{R}^{n_{L_j}}$.
To generate a reconstructed mean vector for the latent field, the decoder takes the form:
\begin{align*}
\v g^D &= (\v z_G^{\top}, \v\phi^{\top})^{\top}, \\
\v h &= \m W_2^D ( \sigma(\m W_1^D \v g^D + \v b_1^D)) + \v b_2^D, \\
\v f_{L_j}^D &= (\v h^{\top}, \v \theta_{L_j}^{\top}, \v\phi^{\top})^{\top}, \quad j \in {1, \ldots, J}, \\
\v \mu_{\v\theta_{L_j}} &= \m W^{L_j}D \v f{L_j}^D + \v b^{L_j}D, \quad j \in {1, \ldots, J}, \\
\hat{\v\theta}{L_j} &\sim \mathcal{N}(\v\mu_{\v\theta_{L_j}}, \gamma\m I), \quad j \in {1, \ldots, J},
\end{align*}
where $\m W_1^D \in \mathbb{R}^{h_D \times (n_G+1)}, \v b_1^D \in \mathbb{R}^{h_D}, \m W_{2}^D\in\mathbb{R}^{h_D\times h_D}, \v b_2^D \in\mathbb{R}^{h_D}$, $\m W^{L_j}D\in\mathbb{R}^{n{L_j}\times (h_D+n_{L_j}+1)}, \v b^{L_j}D\in\mathbb{R}^{n{L_j}}$, and $\gamma>0$ is a fixed scale hyperparameter.

\section{Details for the proper CAR prior}\label{PCAR}
The CAR prior captures the spatial dependence between neighboring regions by conditioning the value of a spatial random effect at a particular location on the values of its neighbors.
In the proper CAR prior, the spatial random effects $\v\phi=(\phi_1,\ldots,\phi_n)^\top$ are assumed to follow a multivariate normal distribution with a covariance matrix that depends on the neighborhood structure of the regions. The joint distribution of the proper CAR prior is given by:
\begin{equation*}
\v\phi \sim \mathcal{N}(\v 0, \sigma^2(\m D - \alpha\m W)^{-1}),
\end{equation*}
where $\m D=\operatorname{diag}(m_1,\ldots,m_n)$ is a diagonal matrix with the number of neighbors for each region, $\m W$ is the adjacency matrix of the neighborhood structure, $\alpha\in (0,1)$ is a spatial dependence parameter, and $\sigma^2$ is a variance parameter. The conditional distributions for each spatial random effect given its neighbors are:
\begin{equation*}
\phi_i \mid \v\phi_{-i} \sim \mathcal{N}\left(\alpha\frac{1}{m_i}\sum_{j\sim i}\phi_j, \frac{\sigma^2}{m_i}\right),
\end{equation*}
where $\v\phi_{-i}$ denotes all elements of $\v\phi$ except $\phi_i$, $m_i$ is the number of neighbors of region $i$, and $j\sim i$ denotes that regions $j$ and $i$ are neighbors. The conditional mean of each random effect is a weighted average of its neighbors, with the weight determined by the spatial dependence parameter $\alpha$. The conditional variance is inversely proportional to the number of neighbors.
The proper CAR prior is proper because $\alpha$ is strictly between 0 and 1. When $\alpha=1$, the proper CAR prior reduces to the intrinsic CAR (ICAR) prior, which is improper because the precision matrix $(\m D - \m W)$ is singular. 
\end{document}